\documentclass[a4paper]{article}

\usepackage{ISCSLP2022}
\usepackage{multirow}

\title{Multilingual Zero Resource Speech Recognition Base on Self-Supervise Pre-Trained Acoustic Models}
\name{Haoyu Wang$^{\dag}$, Wei-Qiang Zhang$^{\dag}$$^*$\thanks{* Corresponding author}, Hongbin Suo$^{\ddag}$, Yulong Wan$^{\ddag}$}
\address{
  $^{\dag}$ Beijing National Research Center for Information Science and Technology \\
  Department of Electronic Engineering, Tsinghua University, Beijing 100084, China \\
  $^{\ddag}$ Data \& AI Engineering System, OPPO, Beijing 100026, China
}
\email{w-hy21@mails.tsinghua.edu.cn, wqzhang@tsinghua.edu.cn, \{suohongbin, wanyulong\}@oppo.com}

\begin{document}
\maketitle
\begin{abstract}
Labeled audio data is insufficient to build satisfying speech recognition systems for most of the languages in the world. There have been some zero-resource methods trying to perform phoneme or word-level speech recognition without labeled audio data of the target language, but the error rate of these methods is usually too high to be applied in real-world scenarios. Recently, the representation ability of self-supervise pre-trained models has been found to be extremely beneficial in zero-resource phoneme recognition. As far as we are concerned, this paper is the first attempt to extend the use of pre-trained models into word-level zero-resource speech recognition. This is done by fine-tuning the pre-trained models on IPA phoneme transcriptions and decoding with a language model trained on extra texts. Experiments on Wav2vec 2.0 and HuBERT models show that this method can achieve less than 20\% word error rate on some languages, and the average error rate on 8 languages is 33.77\%. 

\end{abstract}


\section{Introduction}

For over 7000 languages in the world, only a few of them are served by an Automatic Speech Recognition (ASR) service with enough accuracy. The main barrier on the path to an accurate speech recognition model for these low-resource languages is the insufficiency of labeled training data. The cost to record and transcript a high quality audio corpus is enormous. In contrast, usually, hundreds or thousands of hours of labeled data is needed to train a state-of-art ASR model. Zero-resource speech recognition provides an extreme solution by sharing a well-trained acoustic model with the target language, and can alleviate the reliance on labeled audio data. 

The key point of zero-resource speech recognition methods is the ability to recognize language-independent acoustic units. For instance, International Phonetic Alphabets (IPA) can be used to transcript every language in the world, and are usually used as modeling units in zero-resource methods. Conventionally, an IPA phoneme recognition model needs to be trained in a supervised way with a large amount of labeled training data. Recently, self-supervise pre-trained models provide a new way to model such acoustic units. Some research \cite{baevski2020wav2vec} reveals that the latent speech representations of the Wav2vec 2.0 model are related to phonetic information. Moreover, the HuBERT model is directly trained to predict the hidden units provided by clustering methods \cite{hsu2021hubert}. Pre-trained models have strong generalization ability \cite{erhan2010whydoes}, and are demonstrated to be useful in many low-source tasks \cite{yi2020applying,dai2020lowresourcetts,maimaiti2021lowresourceMT}. Such ability is also helpful for our zero-resource speech recognition task.

Before pre-trained models are introduced, zero-resource methods usually focus on phoneme recognition, and the attributes of phonemes are treated as key points to improve performance. The Universal Phonemic Model (UPM) \cite{li2020zsl} describes all the phonemes by some attributes (such as place or manner of articulation). By comparing the attributes predicted by the model with a pre-defined template, the final phoneme output can be obtained. Allophone adds a projection layer at the end of the model to map global phones to language-specific phonemes \cite{li2020allophone}. Differentiable allophone \cite{yan2021differallo}, a modified version of allophone, uses weighted finite-state transducers to represent this phone-to-phoneme mapping in a differentiable way.

Self-supervised pre-train models bring a significant improvement to zero-resource phoneme recognition tasks.  Gao et al. extract features from Wav2vec 2.0 then use them to train an end-to-end (E2E) model, and get better generalization in unseen languages \cite{gao2021zero}. Xu et al. simply fine-tune the Wav2vec 2.0 model on a multilingual training set and directly test this model on unseen languages \cite{xu2021simple}. This method achieves a state-of-art phoneme error rate (PER) on their test set. This proves that pre-trained models can represent audio in a phonological way without supervised training. Even if the phonemes are missing in the training set, they can still give a relatively accurate approximation.

There are also some works focusing on word-level speech recognition without pre-training methods. Prasad et al. train a Deep Neural Network Hidden Markov Model (DNN-HMM) hybrid model on a rich-resource language which is similar to the target, and replace the language model with another one trained on extra text of the target language \cite{prasad2019building}. This method needs texts, an expert lexicon and phoneme mapping between source and target languages, but these resources are easier to get than labeled audio. 

We believe that the representation ability of pre-trained models can also help in word-level speech recognition. In this paper, we achieve a word error rate (WER) of 13.1\% on the Common Voice Interlingua test set without using any labeled Interlingua training data, and the test on 8 languages shows that our method can get an average WER of 33.77\%. Our method only needs around 5k sentences for each language to train the language models, and the lexicons are also generated by grapheme to phoneme (G2P) tools without expert knowledge. We also optimize the fine-tuning and decoding steps to get better WER. We add word split symbols when fine-tuning, split the diphthongs and triphthongs and extend the lexicon with some possible additional pronounces. Our model significantly outperforms supervised hybrid and E2E models when extremely low text or audio resources are available.

\section{Approach}

\begin{figure}[t]
  \centering
  \includegraphics[width=0.8\linewidth]{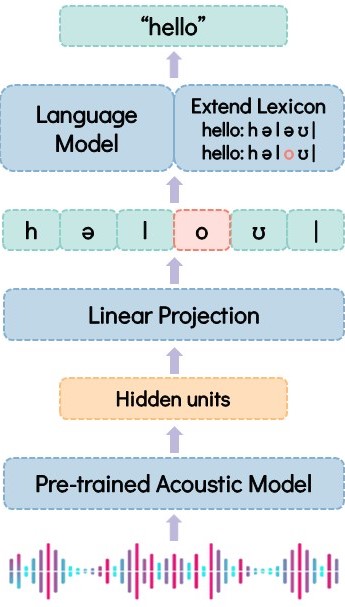}
  \caption{An overview of our method. Notice that the diphthong is split in the model output. Zero-resource recognition may introduce extra mistakes in model output, which can be corrected by the extended lexicon.}
  \label{method}
\end{figure}

To utilize the representation ability of pre-trained models on zero-resource ASR, we still need 2 extra steps. Firstly, these models should be fine-tuned on other languages to set up a projection from self-supervise phone-level units to real IPA phonemes. Secondly, a lexicon and a language model are needed to perform the decoding step. 

For fine-tuning, we want to adjust the pre-trained model to this cross-lingual scenario. As in Figure \ref{method}, we add the word split symbol (``$|$") between the phoneme transcriptions in the training set when fine-tune the pre-trained model, and diphthongs are split in both training and valid set to reduce the out-of-vocabulary (OOV) phonemes when decoding unseen languages.

For decoding, we adjust the lexicons to make our model perform better when recognizing unseen languages. To be specific, when generating the lexicons, additional pronunciations of some words are generated using the Sequitur G2P \cite{bisani2008sequitur} model to make the decoder more possible to generate correct words.

\subsection{Self-supervised pre-trained models}

We select XLSR-53 \cite{conneau2020unsupervised}, HuBERT-large \cite{hsu2021hubert} and Data2vec-large\cite{baevski2022data2vec} as our pre-trained models. XLSR-53 contains a convolutional feature encoder which can map raw audio input into latent speech representations, and a deeper transformer network mapping speech representations into context representations. Besides, during training, speech representations are quantized into a fixed code book by Gumbel-Softmax \cite{jang201gumbel}, and contrastive loss is calculated by selecting the correct quantized representation from $K$ ($K=100$) uniformly sampled distractors \cite{conneau2020unsupervised}.

The HuBERT model is similar to XLSR-53 in structure, but it uses off-line clustering to generate phoneme-level hidden units as ``pseudo-labels", and such pseudo labels are used as the targets of self-supervised training. HuBERT uses cross-entropy loss in optimization. 

The Data2vec-large model is also similar to the above two models in structure, but uses a self-distillation method in training. The model tries to predict the representations of the masked input, and the corresponding ground truth representations are generated by the unmasked version of the input using the same model.  

\subsection{From hidden units to phonemes}

Self-supervise pre-trained models can recognize phoneme-level hidden units, but these units still should be mapped to real IPA phonemes to build the lexicon. This is done by fine-tuning the pre-trained models on a phoneme recognition task. Thanks to the strong representation ability of pre-trained models, we can simply add a linear layer on top of the transformer part and fine-tune them with a training set of dozens of hours, and the languages of the training set are not those in the testing set. 

\subsection{Vowels splitting}

Different languages have different phoneme set, so the existence of out-of-vocabulary phonemes can not be avoided in zero-resource speech recognition. Xu et al. address this by a mapping between OOV phonemes and corresponding substitutions in the training set \cite{xu2021simple} . By our observation, OOV phonemes can be divided into two kinds. Some phonemes contain attributes that are not covered in the training set, and they have to be mapped to some similar ones; the other phonemes, especially diphthongs and triphthongs, can be split into monophthongs that usually can be found in the training set. Base on such an observation, we split all the diphthongs and triphthongs in the transcriptions when training , so the pre-trained model can be adjusted to such a transformation, and can recognize unseen diphthongs and triphthongs as separated monophthongs when testing.

\subsection{Lexicon extending}

Recognizing an unseen language is a great challenge for the acoustic model, and it is obvious that some phonemes may be recognized as similar ones, which sets a barrier to the accuracy of decoding. In our opinion, a flexible lexicon may help in this scenario. Some auto-generated additional pronounces may be a remedy for the mistakes of the acoustic model and help to get the correct decoding results. 

Our lexicon extending method includes two steps. Firstly, we use a generated lexicon to train another G2P model by Sequitur toolkit \cite{bisani2008sequitur}. Secondly, we feed the words in the lexicon back to the new G2P model and get possible additional pronounces of the words. In practice, we noticed that injecting too many wrong pronounces may harm the word error rate, especially for the shorter words, so we only generate lexicon indexes for words longer than 5 characters, and for the remaining parts, only those with confidence over an certain threshold is selected as the final result.

\section{Experimental Setup}

\subsection{Datasets}
We chose Common Voice \cite{ardila2019commonvoice} and Librispeech dataset \cite{panayotov2015librispeech} for training ans testing. Common Voice is an open-source multilingual dataset including over 70 languages, and its total data amount is over 2500 hours. We chose 17 languages with better G2P models from Common Voice version 5.1 for our experiments, in which 9 languages for training and 8 languages for testing. Besides, a subset of Librispeech 100 hours is also combined with the Common Voice training set. For all these languages, we use G2P to transform their transcriptions into phonemes. 

As mentioned above, our zero-resource method is audio-, but it still needs some text of target languages to train a language model and generate a lexicon. For the languages in the testing set, we select at most 10 hours of corresponding training set and used their transcriptions to train the language models. For those languages with less than 10 hours of audio, all the data is used. Table \ref{datset} shows the details of the dataset.

\begin{table}[ht]
  \caption{Details of training and testing set. ``Lang. Code" is the shorten for each language. }
  \label{datset}
  \centering
  \begin{tabular}{llll}
    \toprule
    \textbf{Split} & \textbf{Language} & \textbf{Lang. Code} & \textbf{Total Hours}  \\
    \midrule
    \multirow{10}*{Training} & Czesh & cs & 26 \\
    ~ & Welsh & cy & 83 \\
    ~ & German & de & 692 \\
    ~ & English & en & 100 \\
    ~ & Esperanto & eo & 83 \\
    ~ & Spanish & es & 290 \\
    ~ & French & fr & 554 \\
    ~ & Portuguese & pt & 48 \\
    ~ & Russian & ru & 105 \\
    ~ & Swedish & sv-SE & 10 \\
    \midrule
    \multirow{8}*{Tesing} & Greek & el & 6 \\
    ~ & Basque & eu & 88 \\
    ~ & Interlingua & ia & 5 \\
    ~ & Italian & it & 130 \\
    ~ & Georgian & ka & 3 \\
    ~ & Dutch & nl & 42 \\
    ~ & Polish & pl & 104 \\
    ~ & Romanian & ro & 5 \\
    \bottomrule
  \end{tabular}
\end{table}

\subsection{Baselines}

We use a Kaldi Time-Delay Neural Network (TDNN) hybrid model \cite{povey2011Kaldi} and an Espnet \cite{watanabe2018espnet} end-to-end conformer model as our baselines. The baseline hybrid model is 11-layer TDNN which is trained on 40-dim MFCC features. The baseline E2E model has a 12-layer conformer encoder and a 6-layer transformer decoder. We extract 80-dim Fbank features for E2E training, and use speed perturbing and SpecAugment \cite{park2019specaugment} for data augmentation. 

\subsection{Text pre-processing and normalization}

We use Espeak-ng\footnote{https://github.com/espeak-ng/espeak-ng} to transform words into phoneme transcriptions. To make the transformation more accurate, we only keep five kinds of characters: letters in the target language's alphabet, letters in the English alphabet, spaces, apostrophes and hyphens. We also lowercase all the characters before G2P.  We add ``$|$" as a word splitting symbol when conversion. We also split all the diphthongs and triphthongs, and delete all the stress marks in the final g2p results.

\subsection{Model fine-tuning}

We use the fairseq \cite{ott2019fairseq} open-source pre-trained models. For the fine-tuning of all models, we freeze the CNN feature extractor and and only change the linear layer on the top and the transformer blocks. We set the max update steps at 25k, and the transformer parts are fixed in the first 10k updates. We use the ``tri-stage" learning rate scheduler in fairseq, which linearly increases the learning rate to the settled value in the first 10\% steps, keeps it for the next 40\%, and exponentially decays it to 5\% of the original value in the last 50\% steps. The model is trained on a GTX3090 GPU and it takes around 6 hours to finish the 25k steps. We simply select the last checkpoint as the final result without early stopping. 

\subsection{Lexicon extending and decoding}
Our lexicons are generated and normalized in the same way as the transcriptions. After obtaining the lexicons, we use the Sequitur G2P toolkit to extend the lexicons. 5-order models are trained using the generated lexicons, and 4 possible candidates are generated for each word in the original lexicon. We first delete the words with less than 5 characters, and for the reset, only the ones with top 10\% confidence are selected as the final output and are combined with the original lexicons.

We use Flashlight python binding \cite{kahn2022flashlight} as the beam search decoder. For each target language, a 5-gram Kenlm \cite{heafield2011kenlm} language model is trained on the mentioned extra texts. For the remaining unseen phonemes in the lexicon after vowels splitting, we map them to the nearest ones in the fine-tuning lexicon \cite{xu2021simple}. The decoding beam is set to 50, and we use Sclite\footnote{https://github.com/usnistgov/SCTK} to get the final WER. 

\section{Results}
\subsection{Comparison with the baselines}

\begin{figure}
    \centering
    \includegraphics[width=\linewidth]{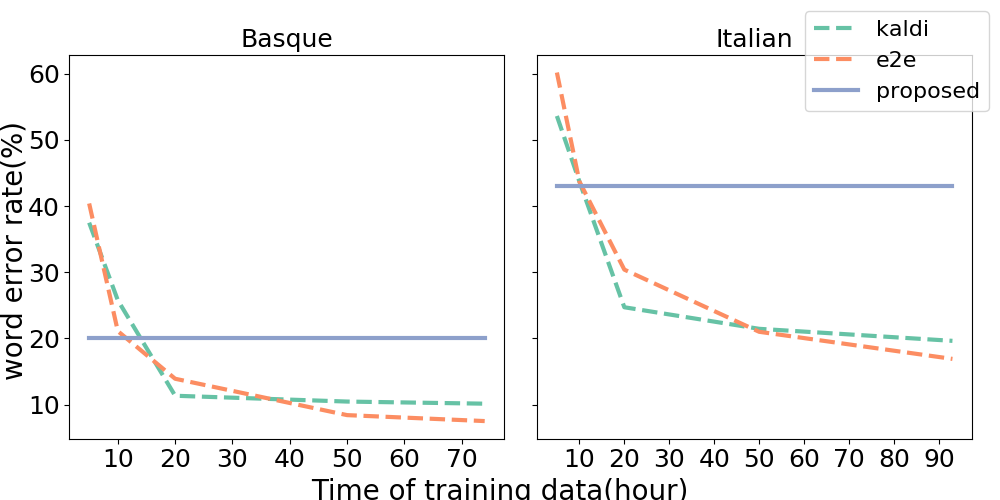}
    \caption{Comparison of the proposed method and the baseline models on different sizes of training data. Our zero-resource method have competitive performance when training data is less than 10 hours.}
    \label{training}
\end{figure}

Firstly, to measure the performance of our method, we compare our models with baseline models which are trained on different sizes of labeled data. We choose Basque and Italian for these experiments, which are relatively rich resource languages in Common Voice. Figure \ref{training} shows that our zero-resource method can have comparable performance with the supervised baseline models trained on around 10 hours of labeled audio. Generally, texts usually cost less to obtain than audio data in practice. As a result, it can be expected that our model may get better results than training a model from scratch in such a scenario. On the other hand, baseline models outperform our method when the training data is over 10 hours. This shows that when enough labeled data can be obtained, it may be better to train the model or try other transfer learning methods. 

\begin{table}[ht]
  \caption{WER of the proposed and baseline models. The baseline models are trained on no more than 10 hours of labeled audio, and the language models are trained on the same amount of texts. Specially, ka, el and eu have no language from the same family in the training set. XLSR, Hu and D2v is for XLSR-53, Hubert-large and Data2vec-large, respectively. }
  \label{baseline}
  \centering
  \begin{tabular}{llllll}
    \toprule
    \multirow{2}*{\textbf{Lang. Code}} & \multicolumn{2}{c}{\textbf{Baseline}} & \multicolumn{3}{c}{\textbf{Proposed}} \\
     & \textbf{E2E} & \textbf{Hybrid} & \textbf{XLSR} & \textbf{Hu} & \textbf{D2v} \\
    \midrule    
    ro & 99.5 & 99.28 & 41.4  & \textbf{41.1} & 49.5 \\ 
    ia & 77.9 & 71.71 & \textbf{13.1}  & 16.1 & 20.6 \\ 
    it & 43.3 & 43.73 & \textbf{43.1}  & 49.3 & 56.1 \\ 
    nl & 51.9 & \textbf{10.26} & 35.0  & 44.1 & 57.1 \\
    pl & 30.5 & \textbf{21.48} & 48.2  & 47.3 & 60.6 \\
    \midrule
    ka & 78.0 & 79.50 & 40.0  & \textbf{33.7} & 45.8 \\
    eu & 21.1 & 25.77 & \textbf{20.1} &  25.4 & 26.0 \\ 
    el & 65.6 & 43.41 & 29.3  & \textbf{26.3} & 38.4 \\
    \midrule
    Avg & 58.47 & 49.39 & \textbf{33.77}  & 35.78 & 44.26 \\
    \bottomrule
  \end{tabular}
\end{table}

We then test our method in low-resource scenarios. In this part, for the baseline models, we split a 10-hour subset for every testing language, and for languages with less than 10 hours of training data, all the data is used. Table \ref{baseline} shows that our methods can reach lower error rates than the supervised baseline models, extremely when the training data is less than 10 hours. 

In Greek, Romanian and Georgian, the HuBERT model outperforms the XLSR-53 model. The HuBERT model is pre-trained only on English unlabeled data, so this result shows that the pre-trained model, especially the HuBERT model, does recognize some language-indepent acoustic units. The Data2vec-large model has higher WER, which may indicate that the Data2vec model is harder to transfer to other languages. This model is pre-trained to predict contextualized representations, which may have large variation between languages. On average, the XLSR-53 model can achieve lower WER, which shows the importance of multi-lingual pre-training. 

Languages in our test set include 7 language families. For some of those languages (for example, Italian and Polish), members from the same languages families can be found in the training set, but for other languages, such as Basque, Georgian or Greek, there is no such correspondence. The results show that whether there exists a similar language in the training set seems to have no influence on the final result. This proves the universality of our method. 


\subsection{Ablation tests}

To prove the effect of our optimization when fine-tuning and decoding, we perform ablation tests of vowel splitting and lexicon extending on XLSR-53 pre-trained model. Table \ref{ablation} shows the details of our test.

\begin{table}[ht]
  \caption{Details of ablation test. ``Proposed" is the fine-tuned XLSR-53 model. Diphthongs are not split when lexicons are not extended.}
  \label{ablation}
  \centering
  \begin{tabular}{lp{1cm}ll}
    \toprule
    \textbf{Lang. Code} & \textbf{Proposed} & \textbf{w/o Splitting} &\textbf{w/o Extending} \\
    \midrule
    eu & \textbf{20.1} & 20.1  & 21.7 \\
    el & \textbf{29.3} & 31.2 & 30.5 \\
    ro & \textbf{41.4} & 41.6 & 43.0\\
    ia & \textbf{13.1} & 14.1 & 14.3\\
    it & \textbf{43.1} & 44.5 & 45.8\\
    nl & \textbf{35.0} & 35.5 & 37.6\\
    pl & \textbf{48.2} & 49.5 & 50.9\\
    ka & \textbf{40.0} & 40.0  & 40.1\\
    \midrule
    Avg. & \textbf{33.77} & 34.56 & 35.48\\
    \bottomrule
  \end{tabular}
\end{table}

Vowel splitting is to alleviate the mismatching of phoneme sets when recognizing unknown languages. In most of our testing languages, diphthongs and triphthongs are the majority of all the out-of-vocabulary phonemes. For instance, over 50\% of OOV phonemes in Italian, Dutch and Romanian are diphthongs or triphthongs, and the improvement after vowel splitting is greater. On the other hand, Basque has lower diphthongs/triphthongs rate, and its improvement is not so obvious. Greek and Dutch are two exceptions. In Greek (el), all the OOV phonemes are diphthongs or triphthongs, but the WER increases after splitting. In Dutch (nl), none of the OOV phonemes are split, but the WER decrease a little. We think these exceptions show the influence of vowel splitting on model fine-tuning. Maybe depended on the way in which vowels are pronounced in different languages, the influence could be positive or negative. 

For every language, final WER decreases after inserting extra pronunciations. Notice that these improvements are realized after deleting the lexicon indexes with less confidence, and without this filtering, the WER usually increases. In our opinion, this phenomenon shows that lexicon extending must be applied with caution and filtering threshold of word length and confidence may need to be adjusted depending on languages.

\section{Conclusion}

In this paper, we apply self-supervise pre-trained model to zero-resource speech recognition. We verify that with a small amount of labeled fine-tune data of other languages and texts of the target language, the pre-trained model can achieve a relatively low word error rate on the test set. We find that splitting the diphthongs and triphthongs when fine-tuning and adding extra pronounces into the lexicon can further decrease the WER. In a scenario where only text resources are available, this method can be used to perform basic speech recognition; and when a small amount of labeled audio can be used, this method can perform better than training a model from scratch. 

\section{Acknowledgements}

This work was supported by the National Natural Science Foundation of China under Grant No. U1836219 and  No. 62276153.


\newpage
\bibliographystyle{IEEEtran}

\bibliography{mybib}

\end{document}